\documentclass[11pt,letterpaper]{article}
\usepackage[hyperref]{acl2017}
\usepackage{times}
\usepackage{latexsym}
\usepackage{url}

\def\aclpaperid{419} 
\aclfinalcopy 

\usepackage{color}
\usepackage{xcolor}
\definecolor{darkgrey}{rgb}{0.2,0.2,0.2}
\definecolor{grey}{rgb}{0.9,0.9,0.9}
\definecolor{darkblue}{rgb}{0.0,0.0,0.5}
\definecolor{darkred}{rgb}{0.5,0.0,0.0}
\definecolor{darkorange}{rgb}{0.5,0.45,0.4}
\definecolor{darkgreen}{rgb}{0.0,0.6,0.0}
\definecolor{darkyellow}{rgb}{1.0,0.65,0.0}
\definecolor{darkergreen}{rgb}{0.0,0.4,0.0}
\definecolor{lightblue}{rgb}{0.8,0.8,1.0}
\definecolor{lightgreen}{rgb}{0.8,1.0,0.8}
\definecolor{lightred}{rgb}{1.0,0.8,0.8}
\definecolor{lightyellow}{rgb}{1.0,1.0,0.8}
\definecolor{lightorange}{rgb}{1.0,0.9,0.8}
\definecolor{lightgrey}{rgb}{0.96,0.97,0.98}
\definecolor{brilliantlavender}{rgb}{0.96, 0.73, 1.0}

\definecolor{mylavender}{HTML}{BD71E1}
\definecolor{darkpurple}{HTML}{531B93}

\usepackage{booktabs}
\usepackage{colortbl}
\usepackage{xcolor}

\usepackage[small]{caption}

\usepackage{adjustbox}
\usepackage{amsmath}
\usepackage{graphicx}

\usepackage{multirow}

\usepackage{microtype}

\newcommand{\word}[1]{{\em #1}}

\newcommand{\mtag}[1]{{\small{\textsf{#1}}}}

\newcommand{\transftxt}[1]{$\text{\sc #1}$}
\newcommand{\transfintabletxt}[1]{\textsc{#1}}

\newcommand{\ptheta}{p_{{\boldsymbol \theta}}}

\long\def\eat#1{\ignorespaces}

\usepackage[disable]{todonotes}   
\newcommand{\note}[4][]{\todo[author=#2,color=#3,size=\scriptsize,fancyline,caption={},#1]{#4}} 
\newcommand{\katha}[2][]{\note[#1]{Katharina}{yellow!40}{#2}}   
   
\usepackage{cleveref}
\crefname{section}{\S}{\S\S}
\Crefname{section}{\S}{\S\S}
\crefname{table}{Tab.}{}
\crefname{table*}{Tab.}{}
\crefname{figure}{Fig.}{}
\crefname{algorithm}{Alg.}{}
\crefname{appendix}{App.}{}
\crefname{equation}{Eq.}{}
\crefformat{section}{\S#2#1#3}  

\title{One-Shot Neural Cross-Lingual Transfer for Paradigm Completion}

\author{Katharina Kann \\ CIS \\  LMU Munich, Germany\\ kann@cis.lmu.de
        \And  Ryan Cotterell \\ Department of Computer Science \\ Johns Hopkins University, USA \\  ryan.cotterell@jhu.edu 
        \And Hinrich Sch\"utze \\ CIS \\  LMU Munich, Germany \\ inquiries@cislmu.org}
          
\date{}

\def\figref#1{Fig.~\ref{fig:#1}}

\def\tabref#1{Tab.~\ref{tab:#1}}
\def\tablabel#1{\label{tab:#1}\label{p:#1}}

\def\secref#1{\S\ref{sec:#1}}

\def\eqref#1{Eq.~\ref{eqn:#1}}

\newcounter{postponednotecounter}
\newcommand{\postponedenotesoff}{\long\gdef\postponedenote##1##2{}}
\newcommand{\postponedenoteson}{\long\gdef\postponedenote##1##2{{
\stepcounter{postponednotecounter}
{\large\bf
\hspace{1cm}\arabic{postponednotecounter} $<<<$ ##1: ##2
$>>>$\hspace{1cm}}}}}
\postponedenoteson
\postponedenotesoff

\newcounter{notecounter}
\newcommand{\enotesoff}{\long\gdef\enote##1##2{}}
\newcommand{\enoteson}{\long\gdef\enote##1##2{{
\stepcounter{notecounter}
{\large\bf
\hspace{1cm}\arabic{notecounter} $<<<$ ##1: ##2
$>>>$\hspace{1cm}}}}}
\enoteson
\enotesoff

\usepackage{draftwatermark}
\SetWatermarkText{Preprint}
\SetWatermarkScale{1.5}

\begin{document}

\maketitle

\begin{abstract}
We present a novel cross-lingual transfer method for paradigm
completion, the task of mapping a lemma to its inflected
forms, using a neural encoder-decoder model, the state of the art for the monolingual task.
We use labeled data from a high-resource language to
increase performance on a low-resource language. In
experiments on 21 language pairs from four different
language families, we obtain up to 58\% higher accuracy than
without transfer and show that even zero-shot and one-shot
learning are possible.  We further find that the degree of
language relatedness strongly influences the ability to
transfer morphological knowledge.
\end{abstract}

\section{Introduction}
Low-resource natural language processing remains an open problem
for many tasks of interest. Furthermore, for most languages in the
world, high-cost linguistic annotation and resource creation are
unlikely to be undertaken in the near future. In the case of
morphology, out of the 7000 currently spoken \cite{Lewis_book_2009} languages,
only about 200 have computer-readable annotations \cite{sylak-glassmankirov2015a} -- although
morphology is easy to annotate compared to syntax and
semantics. {\em Transfer learning} 
is one solution to this problem:
it exploits annotations in a high-resource language
to train a system for a low-resource language. In this work, we
present a method for
cross-lingual transfer of inflectional morphology using an encoder-decoder
recurrent neural network (RNN). This allows for the development
of tools for computational morphology with limited annotated data.

\postponedenote{kk}{I changed this, because the font size was very small and it looked not so nice.}

\begin{table}
\small
\centering
\begin{tabular}{l|llll}
  & \multicolumn{2}{c}{\mtag{PresInd}} & \multicolumn{2}{c}{\mtag{PastInd}}\\
      & \multicolumn{1}{c}{\mtag{Sg}} & \multicolumn{1}{c}{\mtag{Pl}} & \multicolumn{1}{c}{\mtag{Sg}} & \multicolumn{1}{c}{\mtag{Pl}}  \\\hline
  \mtag{1} & \word{sue\~{n}o}         & \word{{so\~{n}amos}} & \word{so\~{n}\'{e}} & \word{so\~{n}amos} \\
  \mtag{2} & \word{sue\~{n}as} & \word{so\~{n}\'{a}is}  & \word{so\~{n}aste} & \word{so\~{n}asteis}  \\
  \mtag{3} & \word{sue\~{n}a}  & \word{sue\~{n}an} & \word{so\~{n}\'{o}} & \word{so\~{n}aron}
\end{tabular}
  \caption{Partial inflection table for the Spanish verb
    \word{so\~{n}ar}\tablabel{partial-paradigm1}}
\end{table}

In morphologically rich languages, individual lexical entries may be
realized as distinct inflections of a single lemma depending on the
syntactic 
context. For example, the \mtag{3SgPresInd} of the English
verbal lemma \word{to bring} is \word{brings}. In many languages,
a lemma can have hundreds of individual forms. Thus, both
generation and analysis of such morphological
inflections are active areas of research in NLP and
morphological processing has been shown to be a boon to several other
down-stream applications, e.g., machine translation
\cite{dyer2008generalizing}, speech recognition
\cite{creutz2007analysis}, parsing \cite{TACL631}, keyword spotting
\cite{narasimhan2014morphological} and word embeddings
\cite{CotterellSE16}, {\em inter alia}. In this work we focus on paradigm
completion, a form of morphological generation that maps a given lemma
to a target inflection, e.g., 
(\word{bring}, \mtag{Past}) $\mapsto$
\word{brought} (with \mtag{Past} being the target tag).

RNN sequence-to-sequence models
\cite{sutskever2014sequence,bahdanau2014neural} are the state
of the art for paradigm completion
\cite{FaruquiTND16,kann-schutze:2016,cotterell-et-al-2016-shared}. However,
these models require a large amount of data to achieve
competitive performance; this makes them
unsuitable for out-of-the-box application to paradigm completion in
the low-resource scenario. To mitigate this, we consider transfer
learning: we train an end-to-end neural system jointly with
limited data from a low-resource language and a larger amount of data
from a high-resource language. This technique allows the model to
apply knowledge distilled from the high-resource training data to the
low-resource language as needed.

We conduct experiments on 21 language pairs from four
language families, emulating a low-resource setting. 
Our results demonstrate successful transfer of morphological
knowledge. We show improvements 
in accuracy and edit distance 
of up to 
58\% (accuracy)
and 4.62 (edit distance) over the same model with only in-domain
language data on the paradigm completion task. We further obtain up to 
44\% (resp.\ 14\%) improvement in accuracy for the one-shot
(resp.\ zero-shot) setting, i.e.,  one (resp.\ zero) in-domain language sample per target tag.
We also show that
the effectiveness
of morphological transfer depends on 
language relatedness,
measured by lexical similarity.

\section{Inflectional Morphology and Paradigm Completion}
Many languages exhibit inflectional morphology, i.e.,
the form of an individual lexical entry mutates to show properties
such as person, number or case. The citation form of a lexical entry
is referred to as the {\bf lemma} and the collection of its possible
inflections as its {\bf
  paradigm}. \tabref{partial-paradigm1} shows an example of a partial
paradigm; we
display several forms for the Spanish verbal lemma \word{so\~{n}ar}. We
may index the entries of a paradigm by a {\bf morphological tag}, e.g.,
the \mtag{2SgPresInd} form \word{sue\~{n}as}
in \tabref{partial-paradigm1}. In generation, the speaker must select
an entry of the paradigm  given 
the form's context. In general, the presence of rich
inflectional morphology is problematic for NLP systems as it greatly
increases the token-type ratio and, thus, word form sparsity.

An important  task in  inflectional morphology
is {\bf paradigm completion}
\cite{durrett2013supervised,ahlberg2014semi,nicolai2015inflection,CotterellPE15,FaruquiTND16}. Its goal is to map a lemma to all individual inflections, e.g.,
$(\text{\word{so\~{n}ar}}, \text{\mtag{1SgPresInd}}) \mapsto \text{\word{sue\~{n}o}}$.
There are good solutions for paradigm
completion when a large amount of annotated training
data is available \cite{cotterell-et-al-2016-shared}.\footnote{The
  SIGMORPHON 2016 shared task
  \cite{cotterell-et-al-2016-shared} on
morphological reinflection,
 a harder generalization of
  paradigm completion, found
that $\geq 98\%$
accuracy can be achieved in many languages with neural
sequence-to-sequence models, improving the state of the art
by 
$10\%$.}  In this work, we address the low-resource
setting, an up to now 
unsolved challenge.

\subsection{Transferring Inflectional Morphology}
In comparison to other NLP annotations, e.g., part-of-speech
(POS) and named entities, morphological inflection does not
lend itself easily to transfer. We can define a universal
set of POS tags \cite{petrov2011universal} or of entity types
(e.g., coarse-grained types like {\em person} and {\em
  location} or fine-grained types \cite{YaghoobzadehS15}),
but inflection is much more language-specific.
It is
infeasible to transfer morphological knowledge from Chinese
to Portuguese as Chinese does not use inflected word
forms. Transferring named entity recognition, however, among
Chinese and European languages works well
\cite{WangM14}. But even transferring inflectional paradigms
from morphologically rich Arabic to Portuguese seems
difficult as the inflections often mark dissimilar
subcategories.  In contrast,
transferring morphological knowledge from Spanish to
Portuguese, two languages with similar conjugations and 89\%
lexical similarity, appears promising. 
Thus, we conjecture that transfer of
inflectional morphology is only viable among {\em related
  languages}.

\subsection{Formalization of the Task}
We  now offer a formal treatment of the cross-lingual
paradigm completion task and develop our notation.
Let $\Sigma_\ell$ be a discrete alphabet for language $\ell$
and let ${\cal T_\ell}$ be a set of morphological tags for $\ell$.
Given a lemma $w_\ell$ in  $\ell$, the morphological paradigm (inflectional table) $\pi$
can be formalized as a set of pairs
\begin{equation}
  \pi(w_\ell) = \Big\{ \big( f_k[w_\ell], t_{k} \big) \Big\}_{k \in T(w_\ell)}
\end{equation}
where $f_k[w_\ell] \in \Sigma_\ell^+$ is an inflected form,
$t_{k} \in {\cal T_\ell}$ is its morphological tag and $T(w_\ell)$ is the
set of slots in the paradigm;
e.g., a  Spanish paradigm is:
\begin{equation*}
  \pi(\text{\tiny \sc so\~{n}ar})\!=\! \Big\{ \big(\text{\footnotesize \word{sue\~{n}o}}, \text{{\mtag{\tiny1SgPresInd}}} \big),\ldots,\big(\text{\footnotesize  \word{so\~{n}aran}}, \text{\mtag{\tiny3PlPastSbj}} \big)\Big\}
\end{equation*}
Paradigm
completion  consists of predicting the entire paradigm
$\pi(w_{\ell})$ given the lemma $w_{\ell}$. 

In cross-lingual paradigm completion, we consider a
\emph{high-resource source language} $\ell_s$ (lots
of training data available) and a \emph{low-resource target language}
$\ell_t$ (little training data available). We denote the source training examples
as ${\cal D}_s$ (with $|{\cal D}_s| = n_s$) and the target
training examples as ${\cal D}_t$ (with $|{\cal D}_t| = n_t$). The goal of
cross-lingual paradigm completion is to populate
paradigms in the low-resource target language with the help of
data from the high-resource source language, using only few in-domain examples.

\section{Cross-Lingual Transfer as Multi-Task Learning}
We describe our probability model for morphological transfer using terminology from
multi-task learning \cite{caruana1998multitask,collobert2011natural}. 
We consider two tasks,  training
a paradigm completor (i) for a high-resource language and (ii)
for a low-resource language. We want
to train jointly so we reap the benefits of having related languages.
Thus, we define the log-likelihood as
\begin{align}
  {\cal L}({\boldsymbol \theta})\!=& \!\!\!\sum_{(k, w_{\ell_t}) \in {\cal D}_t} \!\!\!\!\log \ptheta\left(f_{k}[w_{\ell_t}] \mid w_{\ell_t}, t_{k} \right) \label{eq:ll} \\ 
  & + \!\!\!\sum_{(k, w_{\ell_s}) \in {\cal D}_s}\!\!\!\! \log \ptheta (f_k[w_{\ell_s}] \mid w_{\ell_s}, t_{k}) \nonumber
\end{align}
 where we tie parameters ${\boldsymbol \theta}$ for the two languages together to allow the transfer
of morphological knowledge between languages.
Each probability distribution $\ptheta$ defines a
distribution over all possible realizations of an inflected form,
i.e., a distribution over $\Sigma^*$. \enote{ms}{sigma not
  defined!} For example, consider the
related Romance languages Spanish and French; focusing on one term
from each of the summands in \cref{eq:ll} (the past participle of
the translation of \word{to visit} in each language), we arrive at
\begin{align}
  {\cal L}_{\text{visit}}&({\boldsymbol \theta}) = \log\ptheta(\text{\word{visitado}} \mid \text{\word{visitar}}, \text{\footnotesize \mtag{PastPart}}) \nonumber \\ 
  & + \log\ptheta(\text{\word{visit{\'e}}} \mid \text{\word{visiter}, \text{\footnotesize \mtag{PastPart}}})
\end{align}
Our \emph{cross-lingual} setting
forces both transductions to share part of the parameter vector ${\boldsymbol \theta}$,
to represent morphological regularities between the two languages in a common embedding space
and, thus, to enable morphological transfer. This is no
different from \emph{monolingual} multi-task settings, e.g.,
jointly training
a parser and  tagger for transfer of syntax.

Based on recent advances in neural
transducers, we parameterize each distribution as an encoder-decoder
RNN, as in \cite{kann2016med}. 
In their
setup, the RNN encodes the input and predicts the forms in a {\em
single} language. In contrast, we force the network to predict {\em two} languages.

\subsection{Encoder-Decoder RNN}
\begin{figure}
  \centering
  \includegraphics[width=.75\columnwidth]{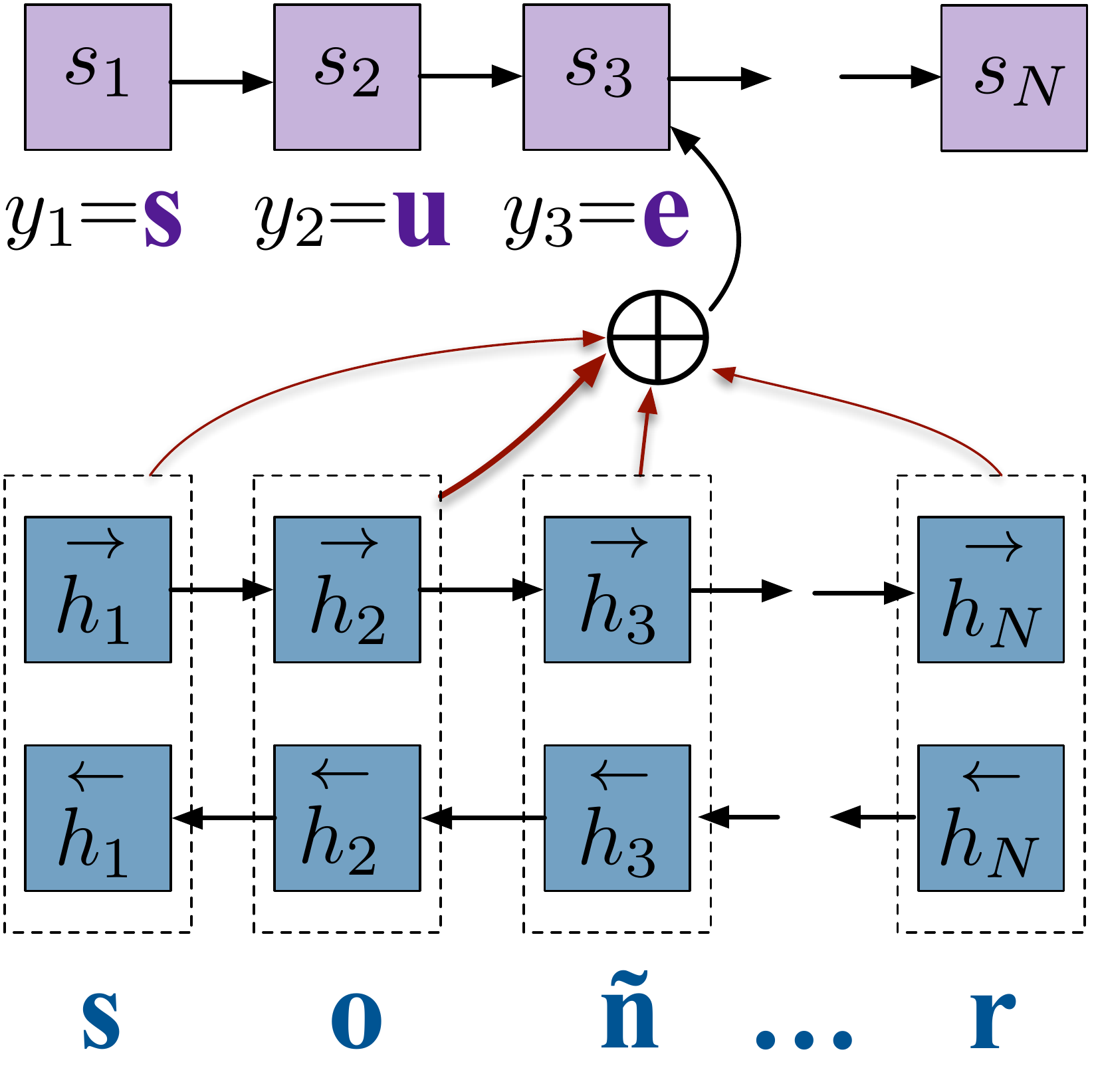}
  \caption{Encoder-decoder RNN
    for paradigm completion. The lemma \word{so\~{n}ar} is mapped to a
    target form (e.g., \word{sue\~{n}a}). For brevity, language and target
    tags are omitted from the input. 
Thickness of 
    red arrows symbolizes the degree to which the model
  attends to the corresponding hidden state of the encoder.}
  \label{fig:model}
\end{figure}

We parameterize the distribution $\ptheta$ as an encoder-decoder gated
RNN with attention \cite{bahdanau2014neural}, the state-of-the-art
solution for the monolingual case \cite{kann2016med}. 
A bidirectional gated RNN encodes the input
sequence \cite{cho2014properties} -- the concatenation of (i) the language tag, 
(ii) the morphological tag of the form to be generated and
(iii) the characters of the input word -- represented by
embeddings. 
The input to the decoder consists of concatenations of
$\overrightarrow{h_{i}}$ and
$\overleftarrow{h_{i}}$,
the forward and backward hidden states of the encoder.
The decoder, a unidirectional RNN,  uses
attention: it
computes a weight for each $h_i$.
Each weight reflects the importance given to that
input position.
Using the attention weights $\alpha_{ij}$, the probability of the output
sequence given the input sequence is:
\begin{align}
\label{eq:2}
  p(y \mid x_1,\ldots, x_{|X|}) 
       &= \prod_{t=1}^{|Y|}g(y_{t-1}, s_t, c_t)
\end{align}
where $y = (y_1, \ldots, y_{|Y|})$ 
is the output sequence (a sequence of $|Y|$ characters),
$x = (x_1, \ldots x_{|X|})$ is the input sequence (a sequence of $|X|$ characters), 
$g$ is a non-linear function, $s_t$ is the hidden state of
the decoder and $c_t$ is the sum of the encoder states $h_{i}$,
weighted by attention weights $\alpha_{i}(s_{t-1})$ which depend on the decoder state:
\begin{equation}
c_t = \sum_{i=1}^{|X|} \alpha_{i}(s_{t-1}) h_{i}
\end{equation}
\figref{model}
shows the encoder-decoder.  See
\newcite{bahdanau2014neural} for further details.

\subsection{Input Format}
Each source form is represented as a sequence of characters; each
character is represented as an embedding.  In the same way, each
source tag is represented as a sequence of subtags, and each subtag is
represented as an embedding. More formally, we define the alphabet
$\Sigma = \cup_{\ell \in L} \Sigma_\ell$ as the set of characters in
the languages in $L$, with $L$ being the set of languages in the given
experiment. Next, we define ${\cal S}$ as the set of subtags that
occur as part of the set of morphological tags $\cal{T}$ $= \cup_{\ell
  \in L} \cal{T}_\ell$; e.g., if \mtag{1\-Sg\-Pres\-Ind} $\in
\cal{T}$, then \mtag{1}, \mtag{Sg}, \mtag{Pres}, \mtag{Ind} $\in
    {\cal S}$. Note that the set of subtags ${\cal S}$ is defined as
    attributes from the {\sc UniMorph} schema \cite{unimorph} and,
    thus, is universal across languages; the schema is derived from
    research in linguistic typology.\footnote{Note that while the
      subtag set is universal, {\em which} subtags a language actually
      uses is language-specific; e.g., Spanish does not mark animacy as Russian
      does. We contrast this with the universal POS set
      \cite{petrov2011universal}, where it is reasonable to expect
      that we see all 17 tags in every language.} The format of the
    input to our system is ${\cal S}^+ \Sigma^+$.  The output format
    is $\Sigma^+$. Both input and output are padded with distinguished
    {\sc bow} and {\sc eow} symbols.

What we have described is  the representation of
\newcite{kann2016med}.  In addition, we preprend a symbol $\lambda \in
L$ to the input string (e.g., $\lambda=$ \mtag{Es}, also represented
by an embedding), so the RNN can handle multiple languages
simultaneously and generalize over them.

\section{Languages and Language Families}
To verify the applicability of our method to a wide range of languages,
we perform experiments on example languages from several different families.

\textbf{Romance languages}, a subfamily of 
Indo-European, are widely spoken, e.g., in
Europe and Latin America. 
Derived from the common ancestor
Vulgar Latin \cite{harris2003romance}, they share
large parts of their lexicon and inflectional morphology; we
expect knowledge among them to be easily transferable.

We experiment on Catalan, French, Italian,
Portuguese and Spanish. \tabref{lexical-similarity} shows that
Spanish -- which takes the role of the low-resource
language in our experiments -- is closely related with the other four, with
Portuguese being most similar. We
hypothesize that the transferability of morphological knowledge
between source and target  corresponds to
the degree of lexical similarity; thus, we expect Portuguese and Catalan to be more
beneficial for Spanish than Italian and French.

The Indo-European \textbf{Slavic language family} has its
origin in eastern-central Europe
\cite{corbett2003slavonic}. 
We experiment on Bulgarian,
Macedonian, Russian and Ukrainian (Cyrillic script)
and on
Czech,
Polish and Slovene (Latin script). 
Macedonian and Ukranian are 
low-resource languages, so we assign them the low-resource
role.
For Romance and for Uralic, we experiment with groups
containing three or four source languages. To arrive at
a comparable experimental setup for Slavic, we run two
experiments, each with three source and one target language:
(i) from Russian, Bulgarian and Czech to Macedonian; and 
(ii) from Russian, Polish and Slovene to
Ukrainian.

We hope that the paradigm
completor learns similar embeddings for, say, the characters
``e'' in Polish and ``$\epsilon$'' in Ukrainian.
Thus, the
use of two scripts in Slavic
allows us to explore transfer across
different alphabets.

\begin{table}
  \centering
\small
\begin{tabular}{l|cccc}
&  \transfintabletxt{pt} & \transfintabletxt{ca} & \transfintabletxt{it} & \transfintabletxt{fr} \\\hline
similarity to \transfintabletxt{es}& 89\%& 85\%& 82\%& 75\%
\end{tabular}
\caption{Lexical similarities for Romance
 \cite{Lewis_book_2009}
\tablabel{lexical-similarity}}
\end{table}

We further consider a non-Indo-European language family, the \textbf{Uralic
languages}.  We experiment on the three most commonly spoken
languages -- Finnish, Estonian and Hungarian \cite{abondolo2015uralic} -- as well as 
Northern Sami, a language used in Northern Scandinavia. While Finnish and Estonian are closely related (both are members of the
Finnic subfamily), Hungarian is a more distant cousin.
Estonian and Northern Sami are 
low-resource languages, so we assign them the low-resource
role, resulting in two groups of experiments:
(i) Finnish, Hungarian and Estonian
to Northern Sami; 
(ii) Finnish, Hungarian and Northern Sami to Estonian.

\textbf{Arabic (baseline)} is a Semitic language (part of the  Afro-Asiatic family
\cite{hetzron2013semitic}) that is
spoken in North Africa, the Arabian Peninsula and other parts of the
Middle East. It is unrelated to all other languages used in this work.
Both in terms of form (new words are mainly built using a templatic system) and categories (it has
tags such as construct state), Arabic is very different.
Thus, we do not expect it to support
morphological knowledge transfer and we use it as a baseline for
all target languages.

\section{Experiments}
We run three experiments on 21 distinct pairings of
languages to show the feasibility of morphological transfer and
analyze our method. We first discuss details common to
all experiments.

\label{subsection:technical_details}
We keep  \textbf{hyperparameters}
during all
experiments (and for all languages) fixed to the following values.
Encoder and decoder RNNs each have 100 hidden units and the size of all subtag, character and language 
embeddings  is 300. For training we use {\sc AdaDelta}
\cite{abs-1212-5701} with minibatch size 20.
All models are trained for 300 epochs.
Following \newcite{LeJH15}, we initialize all weights in the
encoder, decoder and the embeddings except for the GRU weights in the
decoder to the identity matrix. Biases are initialized to zero.

\textbf{Evaluation metrics:}  (i) 1-best
accuracy: the percentage of predictions that match
the true answer exactly; (ii) average edit distance between
prediction and
true answer. 
The two metrics differ in that
accuracy gives no partial credit and incorrect answers may be
drastically different from the annotated form
without incurring additional penalty. In contrast, edit distance gives
partial credit for forms that are closer to the true answer.

\subsection{Exp.\ 1: Transfer Learning for Paradigm Completion}
\label{sec:experiment1}
In this experiment, we investigate to what extent our
model transfers morphological knowledge from a high-resource source language
to a low-resource target language.  
We experimentally answer three questions. (i) Is transfer learning possible for
morphology?  (ii) How much annotated data do we need in the
low-resource target language? (iii) How closely related must the
two languages be to achieve good results?

\begin{figure}
  \includegraphics[width=1.00\columnwidth]{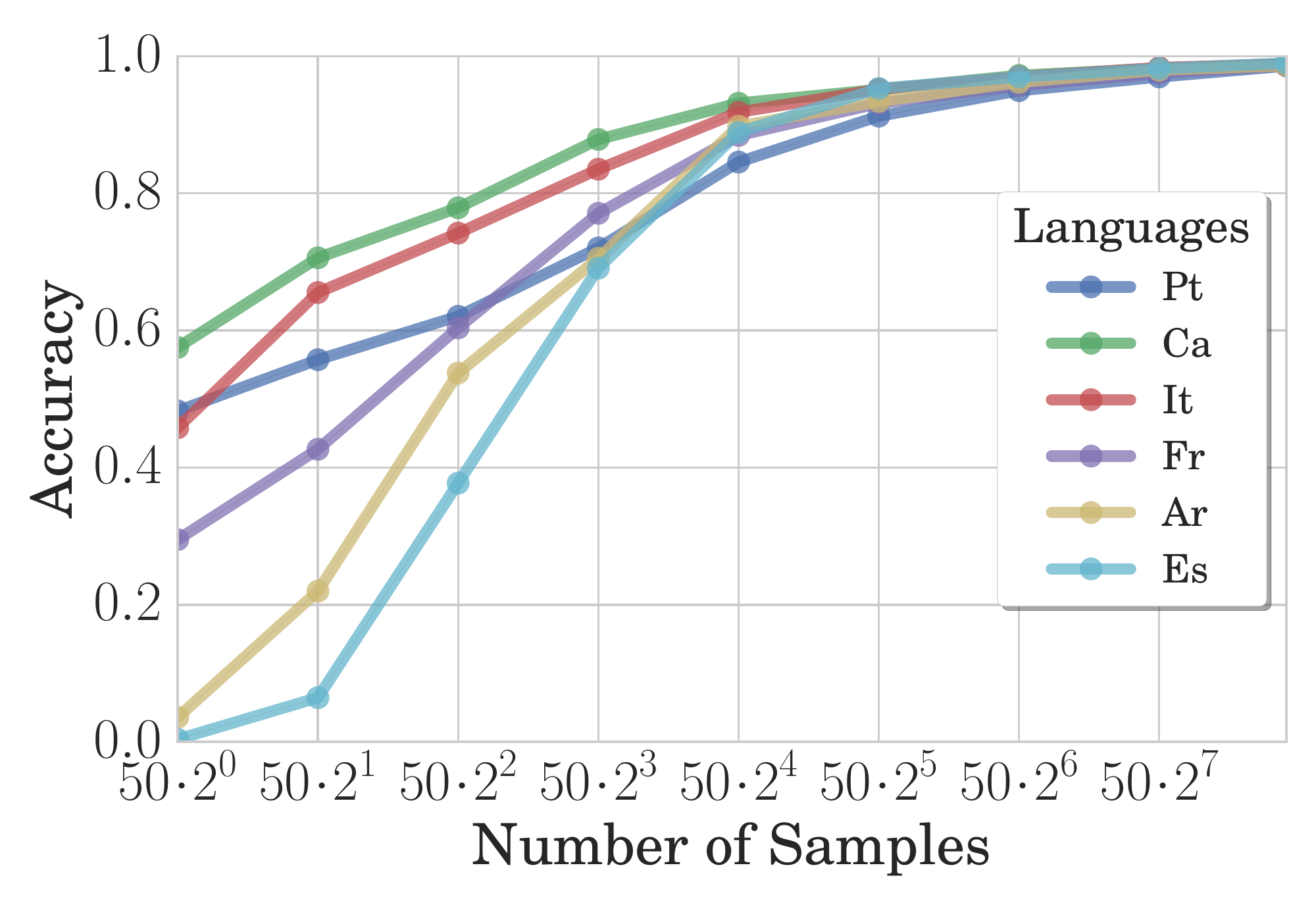}
  \caption{Learning curves showing the accuracy on
    Spanish test   when training on
language $\lambda \in$  \{\transftxt{Pt},
\transftxt{Ca}, \transftxt{It}, \transftxt{Fr}, \transftxt{Ar}, \transftxt{Es}\}. 
Except for $\lambda$=\transftxt{Es}, each model
is
  trained 
on 12,000 samples from $\lambda$  and
``Number of Samples'' (x-axis) of Spanish.
  \label{fig:learning-curve}}
\end{figure}

\def\keytablesep{0.12cm}

\begin{table*}
    \centering
{\scriptsize

    \begin{tabular}{l@{\hspace{0.1cm}}l@{\hspace{\keytablesep}}||@{\hspace{\keytablesep}}c@{\hspace{\keytablesep}}c@{\hspace{\keytablesep}}c@{\hspace{\keytablesep}}c@{\hspace{\keytablesep}}c@{\hspace{\keytablesep}}c@{\hspace{\keytablesep}}|@{\hspace{\keytablesep}}c@{\hspace{\keytablesep}}c@{\hspace{\keytablesep}}c@{\hspace{\keytablesep}}c@{\hspace{\keytablesep}}c@{\hspace{\keytablesep}}|@{\hspace{\keytablesep}}c@{\hspace{\keytablesep}}c@{\hspace{\keytablesep}}c@{\hspace{\keytablesep}}c@{\hspace{\keytablesep}}c@{\hspace{\keytablesep}}|@{\hspace{\keytablesep}}c@{\hspace{\keytablesep}}c@{\hspace{\keytablesep}}c@{\hspace{\keytablesep}}c@{\hspace{\keytablesep}}c@{\hspace{\keytablesep}}|@{\hspace{\keytablesep}}c@{\hspace{\keytablesep}}c@{\hspace{\keytablesep}}c@{\hspace{\keytablesep}}c@{\hspace{\keytablesep}}c@{\hspace{\keytablesep}}c}
&&      
\multicolumn{6}{c@{\hspace{\keytablesep}}|@{\hspace{\keytablesep}}}{Romance}&
\multicolumn{5}{c@{\hspace{\keytablesep}}|@{\hspace{\keytablesep}}}{Slavic I}&
\multicolumn{5}{c@{\hspace{\keytablesep}}|@{\hspace{\keytablesep}}}{Slavic II}&
\multicolumn{5}{c@{\hspace{\keytablesep}}|@{\hspace{\keytablesep}}}{Uralic I}&
\multicolumn{5}{c}{Uralic II}\\
      \multicolumn{2}{c@{\hspace{\keytablesep}}||@{\hspace{\keytablesep}}}{source}  &  \transfintabletxt{0}  &  {\transfintabletxt{ar}}  &  \transfintabletxt{pt}  &  \transfintabletxt{ca}  &  \transfintabletxt{it}  &  \transfintabletxt{fr}  &  \transfintabletxt{0}  &  {\transfintabletxt{ar}}  &  \transfintabletxt{ru}  &  \transfintabletxt{bg}  &  \transfintabletxt{cs}  &  \transfintabletxt{0}  &  {\transfintabletxt{ar}}  &  \transfintabletxt{ru}  &  \transfintabletxt{pl}  &  \transfintabletxt{sl}  &  \transfintabletxt{0}  &  {\transfintabletxt{ar}}  &  \transfintabletxt{fi}  &  \transfintabletxt{hu}  &  \transfintabletxt{et}  &  \transfintabletxt{0}  &  \transfintabletxt{ar}  &  \transfintabletxt{fi}  &  \transfintabletxt{hu}  &   \transfintabletxt{sme}\\
    \multicolumn{2}{c@{\hspace{\keytablesep}}||@{\hspace{\keytablesep}}}{target} & \multicolumn{6}{c@{\hspace{\keytablesep}}|@{\hspace{\keytablesep}}}{{ $\rightarrow$\transfintabletxt{es} } } & \multicolumn{5}{c@{\hspace{\keytablesep}}|@{\hspace{\keytablesep}}}{{ $\rightarrow$\transfintabletxt{mk} } } & \multicolumn{5}{c@{\hspace{\keytablesep}}|@{\hspace{\keytablesep}}}{{ $\rightarrow$\transfintabletxt{uk} } } & \multicolumn{5}{c@{\hspace{\keytablesep}}|@{\hspace{\keytablesep}}}{{ $\rightarrow$\transfintabletxt{sme}} } & \multicolumn{5}{c}{{$\rightarrow$\transfintabletxt{et}} } \\\hline\hline     
    \multirow{2}{*}{\rotatebox{90}{$50$}}  &  acc  &  0.00  &  {0.04}  &  0.48  &  {\bf 0.58}  &  0.46  &  0.29  &  0.00  &  {0.00}  &  0.23  &   \textbf{0.47}  &  0.13  &  0.01  &  {0.01}  &  \textbf{0.47}  &  0.16  &  0.07  &  0.00  &  {0.01}  &  \textbf{0.07}  &  0.05  &  0.03  &  0.02  &  0.01 &  \textbf{0.35}  &  0.21  & 0.17\\
                                             &  ED  &  5.42  &  {4.06}  &  0.85  &  {\bf 0.80}  &  1.15  &  1.82  &  5.71  &  {5.59}  &  1.61  &  \textbf{0.87}  &  2.32  &  5.23  &  {4.80}  &  \textbf{0.77}  &  2.14  &  3.12  &  6.21  &  {5.47}  &  \textbf{2.88}  &  3.46  &  3.71  &  4.50  &  4.51   &  \textbf{1.55}  &  2.19  & 2.60\\\hline 
   \multirow{2}{*}{\rotatebox{90}{$200$}}   &  acc  &  0.38  &  {0.54}  &  0.62  &  {\bf 0.78}  &  0.74  &  0.60  &  0.21  &  {0.40}  &  0.62  &  \textbf{0.77}  &  0.57  &  0.16  &  {0.21}  &  \textbf{0.64}  &  0.55  &  0.50  &  0.13  &  {0.24}  &  {0.26}  &  \textbf{0.28}  &  0.13  &  0.34  &  0.53   &  \textbf{0.74}  &  0.71  & 0.66\\
                                             &  ED  &  1.37  &   {0.87}  &  0.57  &  0.78  &  {\bf 0.44}  &  0.82  &  1.93  &  {1.12}  &  0.68 &  \textbf{0.36}  &  0.72  &   2.09  &  {1.60} &  \textbf{0.49}  &  0.73  &  0.82  &  2.94  &  {1.89} &  {1.78}  &  \textbf{1.61}  &  2.46  &  1.47  &  0.98 &  \textbf{0.41}  &  0.48  & 0.62
  \end{tabular}}
\caption{Accuracy (acc) and edit distance (ED) of
cross-lingual transfer learning for paradigm completion. 
The target language is indicated by
``$\rightarrow$'', e.g., it is Spanish for
``$\rightarrow$\transfintabletxt{es}''. Sources are
indicated in the row ``source''; ``0'' is the monolingual case.
Except for
Estonian, we train on $n_s=12{,}000$ source samples and
$n_t \in \{50, 200\}$ target samples (as indicated by the
row). There are {\em two} baselines in the table. (i) ``0'': no
transfer, i.e., we consider only in-domain data;
(ii) ``\transfintabletxt{ar}'': the Semitic language Arabic is unrelated
to all target languages and functions as a dummy language
that is unlikely to provide  relevant information.  All
languages are denoted using the official codes
(\transfintabletxt{sme}=Northern
Sami). \tablabel{results:exp1_romance}}

\end{table*}

\textbf{Data.}
Based on complete inflection tables from unimorph.org \cite{KIROV16.1077},
we create datasets
as follows.
Each training set consists of 12,000 samples in the
high-resource source language
and $n_t$$\in$$\{$50, 200$\}$ samples in the low-resource
target language. We create target language dev and test sets
of sizes
1600 
and  10,000, respectively.\footnote{For Estonian, we use 
7094 (not 12,000) train and 5000 (not 10,000) test samples
as more data is unavailable.}
For Romance  and Arabic, we create learning curves
for $n_t$$\in$$\{$100, 400, 800, 1600, 3200, 6400, 12000$\}$.  Lemmata
and inflections are randomly selected from all available paradigms.

\katha{Maybe name all pairs here, depending on space.}

\katha{Additionally possible: How to handle different tag sets? Approximations? Mapping!
How to measure "similarity" of languages? What are "features" of "similarity"?}

\katha{A very possible result here would be that we need a tag mapping.}

\textbf{Results and Discussion.}
\tabref{results:exp1_romance} shows the effectiveness of 
transfer learning.  
There are {\em two} baselines. (i) ``0'': no
transfer, i.e., we consider only in-domain data;
(ii) ``\transfintabletxt{ar}'': Arabic, which is unrelated
to all target languages.

With the exception of the 200
sample case of \transftxt{et}$\rightarrow$\transftxt{sme},
cross-lingual transfer is always better than the two
baselines; the maximum improvement is $0.58$ ($0.58$
vs. $0.00$) in accuracy for the 50 sample case of
\transftxt{ca}$\rightarrow$\transftxt{es}.  More closely
related source languages improve performance more than
distant ones. French, 
the Romance language least similar to Spanish,
performs worst for $\rightarrow$\transftxt{es}.
For the target language Macedonian, Bulgarian
provides most benefit.  This can again be explained by
similarity: Bulgarian is closer to Macedonian than the other
languages in this group. The best result for Ukrainian is
\transftxt{ru}$\rightarrow$\transftxt{uk}. Unlike Polish and
Slowenian, Russian is the only language in this group that
uses the same script as Ukrainian, showing the importance of
the alphabet for transfer.  
Still, the results also demonstrate that transfer works
across alphabets (although not as well); this suggests that
similar embeddings for similar characters have been learned.
Finnish is the language that is
closest to Estonian and it again performs best as a source
language for Estonian.  
For Northern Sami, transfer works least well,
probably because  the distance between sources and
target is largest in this case. The distance of the Sami
languages 
from the Finnic (Estonian, Finnish) and Ugric (Hungarian) languages
is much larger than the distances within Romance and within
Slavic.
However, even
for Northern Sami, adding an additional language is still
always beneficial compared to the monolingual baseline.

Learning curves for Romance and Arabic further support our
finding that language similarity is important.  In
\cref{fig:learning-curve}, knowledge is transferred to
Spanish, and a baseline -- a model trained \textit{only} on
Spanish data -- shows the accuracy obtained without any
transfer learning.  Here, Catalan and Italian help the most,
followed by Portuguese, French and, finally, Arabic.  This
corresponds to the order of lexical similarity with Spanish,
except for the performance of Portuguese
(cf.\ \tabref{lexical-similarity}).  A possible explanation
is the potentially confusing overlap of lemmata between the
two languages -- cf.\ discussion in the next subsection.
That the transfer learning setup improves performance for
the unrelated language Arabic as source is at first
surprising. But adding new samples to a small training set
helps prevent overfitting (e.g., rote memorization) even if
the source is a morphologically unrelated language;
effectively acting as a regularizer.\footnote{Following
  \cite{kann2016med} we did not use standard regularizers.
To verify that the effect of Arabic is a regularization
effect, we ran a small monolingual 
experiment  on \transftxt{es} (200 setting)
 with dropout 0.5
\cite{srivastava2014dropout}. The resulting accuracy is
0.57, very similar to the comparable Arabic number of 0.54
in the table.}
This will also be discussed
below.

\textbf{Error Analysis for Romance.}
Even for only 50 Spanish
instances, many inflections are
correctly produced in transfer. 
For, e.g., (\word{criar}, \mtag{3PlFutSbj}) $\mapsto$
\word{criaren}, model outputs are: fr:
\word{criaren}, ca: \word{criaren}, es: \word{crntaron}, it:
\word{criaren}, ar: \word{ecriren}, pt: \word{criaren}
(all correct except for the two baselines).
Many errors involve
accents, e.g., (\word{contrastar}, \mtag{2PlFutInd}) $\mapsto$
\word{contrastar\'{e}is}; model outputs are: fr:
\word{contrastareis}, ca: \word{contrastareis}, es:
\word{conterar\'{i}an}, it: \word{contrastareis}, ar:
\word{contastar\'{i}as}, pt: \word{contrastareis}.  
Some inflections all systems get wrong, mainly because
of erroneously applying the inflectional rules of the source to the target.
Finally, the output of the model trained on Portuguese 
contains a class of errors that are unlike those of other systems.
Example:  (\word{contraatacar},
\mtag{1SgCond}) $\mapsto$ \word{contraatacar\'{i}a} with those
solutions: fr: \word{contratacar\'{i}am}, ca:
\word{contraatacar\'{i}a}, es: \word{concarnar}, it:
\word{contratac\'{e}}, ar: \word{cuntatar\'{i}a} and pt:
\word{contra-atacar\'{i}a}. 
The Portuguese model inserts
``-'' because Portuguese train data
contains \word{contraatacar} and ``-'' appears in its inflected
form.\footnote{To investigate this in more detail we retrain the Portuguese
model with 50 Spanish samples, but exclude all lemmata 
that appear in Spanish train/dev/test, resulting in only 3695 training samples.
Accuracy on test \emph{increases} by $0.09$ despite the
reduced size of the training set.}

An example for the generally improved performance across languages
for 200 Spanish training samples is
(\word{contrastar}, \mtag{2PlIndFut}) $\mapsto$
\word{contrastar\'{e}is}: all models now produce the correct form.

\subsection{Exp.\ 2: Zero-Shot/One-Shot Transfer}
In \secref{experiment1}, we investigated the relationship
between in-domain (target) training set size and performance.
Here, we look at the extreme case of training set
sizes 1 (one-shot) and 0 (zero-shot) for a  tag.
We train our model on \textit{a single} sample for
\textit{half} of the tags appearing in the low-resource language, i.e., if
${\cal T_\ell}$ is the set of morphological tags for the target language, 
train set size is $|{\cal T_\ell}|/2$.
As before, we add 12,000 source samples.

We report 
\textit{one-shot accuracy}
(resp.\ \emph{zero-shot accuracy}), i.e.,
the accuracy for samples with a tag that has been
seen once (resp.\ never) during training.
Note that the model has seen the
\emph{individual subtags} each tag is composed
of.\footnote{It is very unlikely that due to random selection 
a subtag will not be in train; this case did not
occur in our experiments.}

\def\onezeroshotsep{-0.1cm}

\def\partparadigmsep{0.15cm}

\begin{table}
\small
\begin{tabular}{ll ||c@{\hspace{\partparadigmsep}}c@{\hspace{\partparadigmsep}}c@{\hspace{\partparadigmsep}}c@{\hspace{\partparadigmsep}}c@{\hspace{\partparadigmsep}}c}
    & & \transfintabletxt{0} & \transfintabletxt{pt} & \transfintabletxt{ca} & \transfintabletxt{it} & \transfintabletxt{fr} & \transfintabletxt{ar} \\  
    & & \multicolumn{6}{c}{{ $\rightarrow$\transfintabletxt{es}} } \\ \hline\hline
    \multirow{2}{*}{\rotatebox{90}{{\scriptsize \begin{tabular}{l}one\\[\onezeroshotsep] shot\end{tabular}}}}
& acc & 0.00 & \textbf{0.44} & 0.39 & 0.23 & 0.13 & 0.00 \\ 
                                            & ED & 6.26 & \textbf{1.01} & 1.27 & 1.83 & 2.87 & 7.00 \\ \hline
    \multirow{2}{*}{\rotatebox{90}{{\scriptsize \begin{tabular}{l}zero\\[\onezeroshotsep] shot\end{tabular}}}}
    & acc & 0.00 & \textbf{0.14} & 0.08 & 0.01 & 0.02 & 0.00 \\
                                            & ED & 7.18 & \textbf{1.95} & 1.99 & 3.12 & 4.27 & 7.50 
  \end{tabular}
\caption{Results for one-shot and zero-shot transfer learning. Formatting
is the same as for \tabref{results:exp1_romance}. We still use $n_s = 12000$
source  samples. In the one-shot
  (resp.\ zero-shot) case, we observe {\em exactly one form}
  (resp.\ \emph{zero forms}) for each tag in the target
  language at training time.\tablabel{partial_paradigm}
}
\end{table}

\textbf{Data.} Our experimental setup is similar
to \secref{experiment1}: we use the same dev, test  and 
high-resource train sets as before.
However, the low-resource data is created in
the way specified above.
To remove a potentially confounding variable, 
we impose the condition that no two training
samples belong to the same lemma.

\textbf{Results and Discussion.}
\tabref{partial_paradigm} shows that the Spanish and Arabic
systems do not learn anything useful for either
half of the tags. This is not surprising as there is not
enough Spanish data for the system to generalize well and
Arabic does not contribute exploitable information.
The systems trained on French and Italian, in contrast,
get a non-zero accuracy for the zero-shot case as well
as 0.13 and 0.23, respectively, in the one-shot
case. This shows that a single training example is
sometimes sufficient for successful generation although
generalization to tags never observed is rarely possible.
Catalan and Portuguese show the best performance in both
settings; this is intuitive since they are  the languages
closest to
the target  (cf.\ \tabref{lexical-similarity}). In
fact, adding Portuguese to the training data yields an
absolute increase in accuracy of 0.44 (0.44 vs. 0.00) for one-shot
 and 0.14 (0.14 vs. 0.00) for zero-shot with corresponding improvements in
 edit distance.

Overall, this experiment shows that with transfer
learning from a closely related language the performance of
zero-shot morphological generation improves over the monolingual approach, 
and, in the one-shot setting, it is possible to generate the right
form nearly half the time.

\subsection{Exp.\ 3: True Transfer vs. Other Effects}
We would like to separate the effects of regularization that
we saw for Arabic from true transfer.

To this end,
we generate a random cipher (i.e., a function $\gamma:
\Sigma \cup {\cal S}
\mapsto \Sigma \cup {\cal S}$) and apply it to all
word forms and morphological tags of the high-resource train
set; target language data are not changed. 
Ciphering makes it harder to learn
true ``linguistic'' transfer of
morphology. Consider the simplest case of transfer: an
identical mapping in two languages, e.g., (\word{visitar},
\mtag{1SgPresInd}) $\mapsto$ \word{visito} in both
Portuguese and Spanish. If we transform Portuguese using the
cipher $\gamma(\mbox{iostv...}) = \mbox{kltqa...}$, then
\word{visito} becomes \word{aktkql} in Portuguese and its
tag becomes similarly unrecognizable as being identical to
the Spanish tag \mtag{1SgPresInd}. Our intuition is that
ciphering will disrupt
transfer of morphology.\footnote{Note that ciphered input is
  much harder than transfer between two alphabets (Latin/Cyrillic)
because it
  creates ambiguous input. In the example, Spanish ``i'' 
 is totally different from Portuguese ``i'' (which is really
 ``k''), but the model must use the same representation.}
On the other hand, the regularization effect we
observed with Arabic should still be effective.

\textbf{Data.}  We use the Portuguese-Spanish and
Arabic-Spanish data from Experiment 1.  We generate a
random cipher and apply it to morphological tags and word
forms for Portuguese and Arabic.  The
language tags are kept unchanged.  Spanish is also not
changed.  For comparability with 
\tabref{results:exp1_romance},
we use the same dev and test sets as before.

\textbf{Results and Discussion.}
\tabref{results:exp3} shows that
performance of
\transfintabletxt{pt}$\rightarrow$\transfintabletxt{es}
drops a lot: from 0.48 to 0.09 for 50 samples and from 0.62
to 0.54 for 200 samples. This is because there are no overt
similarities between the two languages left after applying
the cipher, e.g., the two previously identical forms
\word{visito} are now different.

The impact of ciphering on
\transfintabletxt{ar}$\rightarrow$\transfintabletxt{es}
  varies: slightly improved in one case (0.54 vs.\ 0.56),
slightly worse in three cases. We also apply the
cipher to the tags and Arabic and Spanish share subtags,
e.g., \mtag{Sg}. 
Just the knowledge that something is a subtag is helpful
because subtags must not be generated as part of the output.
We can explain the tendency of ciphering to decrease
performance on
\transfintabletxt{ar}$\rightarrow$\transfintabletxt{es} 
by the ``masking'' of common subtags.

For 200 samples and ciphering, there is no clear  difference in
performance between Portuguese and Arabic.
However, for 50 samples and ciphering, Portuguese 
(0.09) seems to perform better than Arabic (0.02) in accuracy.
Portuguese uses suffixation for inflection whereas Arabic is
templatic and inflectional changes are not limited to the
end of the word. This difference is not affected by
ciphering. 
Perhaps even ciphered Portugese lets the model learn better that the beginnings
of words just need to be copied. 
For 200 samples, the
Spanish dataset may be large enough, so that ciphered
Portuguese no longer helps in this regard.

Comparing no transfer
with transfer from a ciphered language to Spanish, we
see large performance gains, at least for the 200 sample
case: 0.38
(\transfintabletxt{0}$\rightarrow$\transfintabletxt{es})
vs.\
0.54
(\transfintabletxt{pt}$\rightarrow$\transfintabletxt{es})
and 0.56
(\transfintabletxt{ar}$\rightarrow$\transfintabletxt{es}). 
This
is evidence that our conjecture is correct that the baseline
Arabic mainly acts as a regularizer that prevents the model
from memorizing the training set and therefore improves
performance. So performance improves even though no true
transfer of morphological knowledge takes place. 

\begin{table}
\small
\begin{tabular}{ll ||c|cc|cc}
& & 0$\rightarrow$\transfintabletxt{es}& \multicolumn{2}{c}{{ \transfintabletxt{pt}$\rightarrow$\transfintabletxt{es}} } & \multicolumn{2}{c}{{ \transfintabletxt{ar}$\rightarrow$\transfintabletxt{es}} } \\ 
    & &  & orig & ciph & orig & ciph \\  \hline\hline
     \multirow{2}{*}{\rotatebox{90}{$50$}} & acc & 0.00 & 0.48 & 0.09 & 0.04 & 0.02 \\
                                            & ED & 5.42 & 0.85 & 3.25 & 4.06 & 4.62 \\ \hline
   \multirow{2}{*}{\rotatebox{90}{$200$}}  & acc & 0.38 & 0.62 & 0.54 & 0.54 & 0.56 \\ 
                                            & ED & 1.37 & 0.57 & 0.95 & 0.87 & 0.93 \\
  \end{tabular}
\caption{Results for ciphering. 
``0$\rightarrow$\transfintabletxt{es}''
and ``orig'' are original results, copied 
  from \tabref{results:exp1_romance}; ``ciph'' is the result after
  the cipher has been applied.
\tablabel{results:exp3}}
\end{table}

\section{Related Work}
\textbf{Cross-lingual transfer learning}  
has been used for many
tasks: automatic speech recognition
\cite{huang2013cross}, parsing
\cite{CohenDS11,sogaard:2011:ACL-HLT20112,Naseem1}, entity
recognition \cite{MengqiuWang2014} and machine translation
\cite{JohnsonSLKWCTVW16,ha2016toward}.
One straightforward method
is to translate 
datasets
and then train a monolingual model
\cite{fortuna2005use,olsson2005cross}. Also, aligned corpora
have been used to project information from annotations in one language
to another \cite{yarowsky2001inducing,pado2005cross}.
The drawback is that
machine translation errors cause errors in the target.
Therefore, alternative methods have
been proposed, e.g., to port a model trained on the source
language to the target language \cite{ShiMT10}.

In the realm of morphology, \newcite{buys-botha:2016:P16-1} recently adapted methods
for the training of POS taggers to learn weakly supervised morphological taggers
with the help of parallel text. 
Snyder and Barzilay (2008a, 2008b)
\nocite{DBLP:conf/aaai/SnyderB08} 
\nocite{snyder-barzilay:2008:ACLMain} 
developed a non-parametric Bayesian model for morphological segmentation. They performed identification of
cross-lingual abstract morphemes and segmentation
simultaneously and reported, similar to us, best results for related languages.

Work on \textbf{paradigm completion} has recently been encouraged by the
SIGMORPHON 2016 shared task on morphological reinflection
\cite{cotterell-et-al-2016-shared}.  Some work first applies an unsupervised
alignment model to source and target string pairs 
and then learns a string-to-string mapping
\cite{durrett2013supervised,nicolai2015inflection}, using, e.g.,
a semi-Markov conditional random field
\cite{sarawagi2004semi}. 
Encoder-decoder RNNs
\cite{aharoni2016improving,faruqui2015morphological,kann2016med},
a method which our work further develops for the
cross-lingual scenario, define the current state of the art.

\textbf{Encoder-decoder RNNs} were developed in parallel by
\newcite{cho2014properties} and
\newcite{sutskever2014sequence} for machine translation and
extended by \newcite{bahdanau2014neural} with an attention
mechanism, supporting better generalization. They have been
applied to NLP tasks like speech recognition
\cite{graves2005framewise,graves2013speech}, parsing
\cite{vinyals2015grammar} and segmentation
\cite{kann2016neural}.  More recently, a number of papers
have used encoder-decoder RNNs in \emph{multitask and transfer
learning settings}; this is mainly work in machine
translation (MT):
\cite{dong15multitask,zoph2016multi,chu2017,JohnsonSLKWCTVW16,luong2016iclr_multi,firat16multiway,ha2016toward},
     {\em inter alia}.
Each of these papers has both similarities and differences
with our approach. (i) Most train several distinct models
whereas we train a \emph{single model} on input augmented with an
explicit encoding of the language (similar to \cite{JohnsonSLKWCTVW16}). 
(ii) Let $k$ and $m$ be the number of different input and output
languages.  We address the case $k \in \{1,2\}$ and $m=k$.
Other work has addressed cases with $k>2$ or $m>2$; this would
be an interesting avenue of future research for paradigm
completion. 
(iii) Whereas training RNNs in MT is hard, we only
experienced one difficult issue in our experiments (due to
the low-resource setting):
regularization.
(iv) Some work is word- or subword-based, our
work is character-based. The same way that similar word
embeddings are learned for the inputs \emph{cow} and
\emph{vache} (French for ``cow'') in MT, we expect similar
embeddings to be learned for similar Cyrillic/Latin characters. (v)
Similar to work in MT, we show that zero-shot (and, by extension,
one-shot) learning is possible.

\cite{ha2016toward} 
(which was developed in parallel to our transfer model although we
did not prepublish our paper on arxiv)
is most similar to
our work. Whereas 
\newcite{ha2016toward} address MT, we focus
on the task of paradigm completion in
low-resource settings and establish the state of the art
for this problem.

\section{Conclusion}
We presented a cross-lingual transfer learning
method for paradigm completion, based on an RNN encoder-decoder model.
Our experiments showed that information from a high-resource language can be
leveraged for paradigm completion in a related low-resource language.
Our analysis showed that the degree to which the source language data helps
for a certain target language depends on their relatedness. 
Our method led to significant
improvements in settings with limited training data -- up to 58\% absolute improvement in
accuracy -- and, thus, enables
the use of 
state-of-the-art models for
paradigm completion in low-resource languages.

\postponedenote{hs}{
acknowledgments

Sylak-Glassman and Christo Kirov went out of their
way to create the resources that this paper is based on

}

\bibliography{crossling}
\bibliographystyle{acl_natbib}

\end{document}